\documentclass{article}

\usepackage{arxiv}

\usepackage{amsmath,amsfonts,bm}

\def\eqref#1{equation~\ref{#1}}

\def\1{\bm{1}}

\DeclareMathAlphabet{\mathsfit}{\encodingdefault}{\sfdefault}{m}{sl}
\SetMathAlphabet{\mathsfit}{bold}{\encodingdefault}{\sfdefault}{bx}{n}

\usepackage[utf8]{inputenc} 
\usepackage[T1]{fontenc}    
\usepackage{hyperref}       
\usepackage{url}            
\usepackage{booktabs}       
\usepackage{amsfonts}       
\usepackage{nicefrac}       
\usepackage{microtype}      
\usepackage{lipsum}		
\usepackage{graphicx}
\usepackage{natbib}
\usepackage{doi}

\usepackage{mathtools}
\newcommand{\norm}[1]{\left\lVert#1\right\rVert}

\title{Molecular Graph Generation via Geometric Scattering}

\author{ 
\href{https://orcid.org/0000-0001-8068-3101}{\includegraphics[scale=0.06]{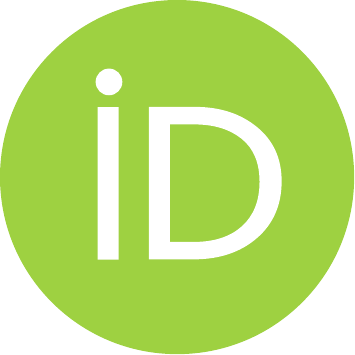}\hspace{1mm}Dhananjay Bhaskar}\\
	Department of Genetics\\
	Yale University\\
	New Haven, CT 06511, USA \\
	\texttt{dhananjay.bhaskar@yale.edu} \\
	\And
	\href{https://orcid.org/0000-0003-4521-4187}{\includegraphics[scale=0.06]{orcid.pdf}\hspace{1mm}Jackson D. Grady} \\
	Department of Computer Science \\
	Yale University \\
	New Haven, CT 06511, USA \\
	\texttt{jackson.grady@yale.edu} \\
	\AND
	 \href{https://orcid.org/0000-0001-9829-4016}{\includegraphics[scale=0.06]{orcid.pdf}\hspace{1mm}Michael A. Perlmutter} \\
	 Department of Mathematics \\
	 UCLA \\
	 Los Angeles, CA 90095, USA \\
	 \texttt{perlmutter@math.ucla.edu} \\
	 \And
	 \href{https://orcid.org/0000-0001-5823-1985}{\includegraphics[scale=0.06]{orcid.pdf}\hspace{1mm}Smita Krishnaswamy} \\
	 Department of Computer Science \\
     Department of Genetics \\
     Yale University \\
     New Haven, CT 06511, USA \\
	 \texttt{smita.krishnaswamy@yale.edu} \\
}

\date{}

\renewcommand{\headeright}{}
\renewcommand{\undertitle}{}

\renewcommand{\shorttitle}{Molecular Graph Generation via Geometric Scattering}

\hypersetup{
pdftitle={Molecular Graph Generation via Geometric Scattering},
pdfsubject={cs.LG, stat.ML},
pdfauthor={Dhananjay Bhaskar, Jackson D. Grady, Michael A. Perlmutter, Smita Krishnaswamy},
pdfkeywords={Drug Discovery, Graph Generation, Geometric Scattering, Autoencoder},
}

\begin{document}
\maketitle

\begin{abstract}
Graph neural networks (GNNs) have been used extensively for addressing problems in drug design and discovery. Both ligand and target molecules are represented as graphs with node and edge features encoding information about atomic elements and bonds respectively. Although existing deep learning models perform remarkably well at predicting physicochemical properties and binding affinities, the generation of new molecules with optimized properties remains challenging. Inherently, most GNNs perform poorly in whole-graph representation due to the limitations of the message-passing paradigm. Furthermore, step-by-step graph generation frameworks that use reinforcement learning or other sequential processing can be slow and result in a high proportion of invalid molecules with substantial post-processing needed in order to satisfy the principles of stoichiometry. To address these issues, we propose a representation-first approach to molecular graph generation. We  guide the latent representation of an autoencoder by capturing graph structure information with the geometric scattering transform and apply penalties that structure the representation also by molecular properties. We show that this highly structured latent space can be directly used for molecular graph generation by the use of a GAN.  We demonstrate that our architecture learns meaningful representations of drug datasets and provides a platform for goal-directed drug synthesis.
\end{abstract}

\keywords{Drug Discovery \and Molecular Graph Generation \and Geometric Scattering}

\section{Introduction}
Recently there has been a great deal of interest in developing neural networks for graph-structured data. However, the vast majority of the literature on graph  neural networks and variants has focused on node embeddings and node classification. Because of this, most GNNs are designed to produce features for nodes based on neighborhood aggregation, which gives only localized information.  A smaller subset of this literature has focused on graph classification. These networks have recently proposed using longer-range information via skip connections or attention mechanisms \citep{abu2019mixhop, yun2019graph}. However, there has been relatively little attention given to graph generation, with the most common approaches being somewhat cumbersome methods of sequential or reinforcement-learning based generation. Here, we seek to remedy this by focusing on a representation-first approach to graph generation that we call GRASSY (GRAph Scattering SYnthesis network). GRASSY focuses on producing a latent space embedding of molecular graphs that is organized both by molecular structure and physicochemical properties. It is then trained adversarially to generate molecules with desirable properties directly from this latent space. 

The GRASSY framework makes use of the geometric scattering transform in order to learn a rich representation of the graph. This transform discretizes the original scattering transform of \citet{mallat2012group} and uses multiscale graph diffusion wavelets to form globally-contextualized descriptions of each graph. Notably, the version of the scattering transform we use here collects statistical moments of multi-scaled diffusion wavelet coefficients through global summation.  Therefore, the representation it produces is fully permutation-invariant, and the number of moments we obtain does not depend on the size of the original graph. 

After computing the scattering transform, GRASSY reduces the dimensionality of the resulting space with an autoencoder which is penalized by a reconstruction penalty and a property-prediction penalty. This yields a highly structured latent space, which we may sample from in order to generate scattering coefficients corresponding to molecular graphs with desirable properties. To complete the graph generation process, GRASSY utilizes an adversarial framework which produces molecular graphs directly from the latent space. Importantly, we note that GRASSY does not use a sequential process or reinforcement-learning-based method for molecular synthesis but rather immediately generates molecules based on the organization of the latent space. 

In summary, the key components of GRASSY include: 
\begin{itemize}
\item A geometric scattering network to generate multiscale descriptions of graphs 
\item A regularized autoencoder for producing  representations of molecules in a structured latent space
\item An adversarial molecular generation network.   
\end{itemize}

We show the utility of GRASSY on two datasets described in detail in Section \ref{results}. 1) ZINC, a dataset of drug-like molecules with several properties of each molecule, 2) BindingDB, a database of drug-target interactions. We show that GRASSY learns the latent space of several tranches of ZINC and produces drug-like molecules. We also show that it can learn molecules with binding affinities to specific targets and generate molecules in this space as well.

\section{Related work on Molecule Generation via Graph Neural Nets}

A variety of notable approaches have been made to tackle the difficult problem of molecular graph generation. GraphAF \citep{shi_graphaf_2020} leverages a flow-based autoregressive model, which formulates graphs as a sequential decision process and then applies reinforcement learning to generate graphs with specified properties. MolGAN \citep{de_cao_molgan_2018} takes a different approach, using a GAN-based architecture. From normally distributed samples, it generates graphs using a discriminator which is based  on the Graph Convolutional Network (GCN) of \citet{kipf_semi-supervised_2017}. This discriminator also penalizes graphs based on a reward network, which encourages the generated graphs to have specific, desired properties. Similarly, LGGAN  \citep{fan_labeled_2021} uses a GAN-based architecture, and also uses a GCN as a means to discriminate between real and fake graphs. In a different approach, \cite{gomez-bombarelli_automatic_2018}, generate graphs by creating an auto-encoder that encodes and decodes SMILES strings, which allowed them to sample points from the latent space and generate new SMILES strings.

\section{Preliminaries and Background}

\subsection{Graph diffusion}

The geometric scattering transform introduced in the following subsection uses multi-scale diffusion wavelets which are inspired, in part, by methods from high-dimensional data analysis \citep{coifman_diffusion_2006}. 
In many applications, one is given a data set $\{\mathbf{x}_i\}_{i=1}^n$ contained in very high-dimensional Euclidean space $\mathbb{R}^N$. The excessively large dimension of such a data set makes it hard to analyze and expensive to store or process. Fortunately, in many applications, the data has an intrinsic lower-dimensional structure and can be modeled as lying along a $d$-dimensional Riemannian manifold for some $d\ll N$. To exploit this low-dimensional structure, one aims to construct a graph whose geometry models the underlying manifold. A popular method for doing this is to let the data points $\mathbf{x}_i$ be the vertices of the graph and to define the affinity between vertices using a predefined kernel function. A common choice is to define a weighted edge between vertices $i$ and $j$, $i\neq j$ with weight given by
$W_{i,j} \coloneqq e^{-\|\mathbf{x}_i - \mathbf{x}_j\|_2^2 / \epsilon}$ for a suitably chosen parameter $\epsilon$. 
One may view the $W$ as the weighted adjacency matrix of a weighted graph $G=(V,E,W)$. (Here, we set $W_{i,i}=0$ so that the graph has no self-loops.) 

The matrix $W$ encodes \emph{local} information about the graph $G$. In order to capture information about the graphs global geometry, we introduce a lazy random walk matrix
\begin{equation}\label{eqn: P}
    P \coloneqq \frac{1}{2}(I_n + WD^{-1}), 
\end{equation}
where $I_n$ is the $n\times n$ identity matrix and $D$ is the diagonal degree matrix whose non-zero entries are defined by $D_{i,i}=\sum_{j=1}^nW_{i,j}$.
Raising the matrix $P$ to different powers captures information about the graph at various degrees of resolution. For example, $P^2$ encodes information within two-step neighborhoods of each vertex whereas $P$ raised to a very large power captures global averages. The popular diffusion maps algorithm, introduced in \citet{Diffusionmaps}, is based on the eigendecomposition of this matrix $P$ and its powers. It has been shown to be useful for many applications including data visualization \citep{phate}, denoising \citep{van2017magic}, and  trajectory analysis \citep{haghverdi2016diffusion}.

\subsection{Geometric Scattering}

The scattering transform, originally introduced in \cite{mallat2012group} for Euclidean data is a wavelet-based, feed-forward network which produces a latent-space representation of an input via an alternating sequence of convolutions and nonlinearities. The architecture of the scattering transform is similar to a convolutional neural network but uses predesigned filters rather than filters learned from training data. Inspired by the rise of graph neural networks, several works \citep{zou2020graph,gama2018diffusion,pmlr-v97-gao19e} have adapted the scattering transform to the graph setting and analyzed the stability of the resulting networks \citep{perlmutter_understanding_2019, gama2019stability}. Similar to \cite{mallat2012group}, the original formulations of the graph scattering transforms were fully designed networks using dyadic wavelets. However, subsequent work has incorporated learning via cross-channel convolutions \citep{min2020scattering}, attention mechanisms \citep{min2021geometric}, or by replacing dyadic-scales with scales learned from data \citep{tong_data-driven_2021}. Most closely to our work, \citep{zou2019encoding} and \citep{castro_uncovering_2020} have shown that the scattering transform can be incorporated into an encoder-decoder type network. Here we utilize this framework for molecular graph generation. 

\subsection{Scattering moments for graph structure representation}

The primary focus of \cite{pmlr-v97-gao19e} was applying the graph scattering transform to graph classification. There, the authors construct the scattering transform as an alternating cascade of wavelet transforms and pointwise nonlinearities. In particular, they use   diffusion wavelets constructed from the matrix $P$ defined in \eqref{eqn: P} raised to dydadic powers. Specifically, for $J\geq 1,$ they define
\begin{equation} \label{eqn: wavelet def}
 \Psi_0 \coloneqq I_n - P \, , \quad \Psi_j \coloneqq P^{2^{j-1}} - P^{2^j} = P^{2^{j-1}} ( I_n - P^{2^{j-1}}) \, , \quad j \geq 1 \, .
\end{equation}
Different powers of $P$ capture information about the graph at different scales. In particular, because of the larger values of $j$, the scattering transform is able to capture both local and long-range information in a single layer. This is in stark contrast to many popular graph neural networks such as GCN  \citep{kipf_semi-supervised_2017} in which the convolutions are purely local averaging operations.

Given these diffusion wavelets, first and second-order scattering coefficients are defined by
\begin{equation} \label{eqn: scat coeff}
 S\mathbf{x}(j, q) \coloneqq \sum_{l=1}^n |\mathbf{\Psi}_j\mathbf{x}(v_l)|^q,\quad \text{and}\quad S\mathbf{x}(j, j', q) \coloneqq \sum_{l=1}^n |\mathbf{\Psi}_{j'}|\mathbf{\Psi}_j\mathbf{x}(v_l)||^q.
\end{equation}
for $1\leq j,j'\leq J$ and $1\leq q\leq Q.$ The authors also use zeroth-order coefficients which are simply defined by $S(q)\coloneqq \|\mathbf{x}\|_q^q.$
The graph scattering transform is then defined by concatenating all of these coefficients:
\begin{equation} \label{eqn: scat transform}
S\mathbf{x}=\{ S\mathbf{x}(q), S\mathbf{x}(j,q), S\mathbf{x}(j,j',q), 1\leq j,j'\leq J, 1\leq q\leq Q\}.
\end{equation}
We note that since these coefficients are defined via a global summation they are fully invariant to permutations of the vertices. Moreover, the number of coefficients does not depend on the size of the graph. This allows one to apply the graph scattering transform to data sets consisting of many graphs where each graph may have a different number of vertices.

\subsection{Geometric Scattering Autoencoder}

In \cite{castro_uncovering_2020}, the authors proposed to autoencoder scattering transforms of folds of an single non-coding RNA molecules in order to infer potential folding trajectories from latent space visualization. They used a fixed three-scale scattering transform to obtain multi-level wavelet coefficients at each node which they concatenated and used as input to a variational autoencoder network. They showed that this framework called {\em Graph Scattering Autoencoder (GSAE)} is able to preserve similarity structure in the folds, as well as fold energy, in a low-dimensional embedding.  The authors also proposed a scattering inversion network (SIN), which is an autoencoder whose middle layer is an adjacency matrix. The SIN purports to go back from the scattering coefficients to an adjacency matrix. However, the SIN network only produces meaningful results with very strong initializations or guided searches around a small area of the graph space, mainly used to interpolate between two adjacent folds of the same RNA molecule. 

Unlike GSAE, our primary aim is to generate novel molecules of different sizes from the embedded space of a neural network trained on a variety of drug-like molecules. For this purpose, we build upon and expand GSAE in several ways. 1) We use scattering moments, i.e., aggregation of scattering transform coefficients which allows us to handle graphs of different sizes. 2) We use a GAN operates directly on the learned latent space. This molecular generation network is trained with an adversarial loss as well as a latent interpolation loss that obviates the need for both the SIN and the variational training in GSAE.  3) We utilize a differentiable scattering transform \cite{tong_data-driven_2021} such that the scales of scattering are not fixed, but instead learned from the data. In the next section, we detail features of our network which we call GRASSY.

\section{Methods}
\label{methods}

\subsection{Problem Formulation}\label{sec: formulation}
Our main goal is to find a latent space representation that is smooth with respect to various molecular properties as well as graph edit distance and to use such a representation to generate well-formed molecular graphs. More formally, given molecular graphs $G_1=\{E_1, V_1\}, G_2=\{E_2, V_2\}$, with molecular properties $p(G_1)$, $p(G_2)$, we seek a embedding $f$, and a pseudo-inversion $f'$ such that:

\begin{itemize}
    \item If $\|f(G_1)-f(G_2)\|_2\leq \epsilon_e$, then $\text{dist}(G_1, G_2)<\nu$,  i.e., close graphs have close embedded distances. 
    \item If $\|f(G_1)-f(G_2)\|_2\leq \epsilon_p$, then $\text{dist}(p(G_1), p(G_2))<\mu$, i.e., close graphs have similar molecular properties. 
    \item $\text{dist}(f'(f(G_1)+e), G_1) \leq \epsilon_g$, where $e \sim N(0,\sigma)$, i.e. a graph generated by pseudo-inversion after perturbation of embedding $f(G_1)$ in the latent space, is similar to the original graph. 
    \item $\text{dist}(p(f'(f(G_1)+e)), p(G_1)) \leq \epsilon$, where $e \sim N(0,\sigma)$, i.e., a graph generated by pseudo-inversion after perturbation of embedding $f(G_1)$ in the latent space has similar molecular properties as the original graph. 
\end{itemize}

Typically, molecular graphs are not scale-free and have complex connectivity structure. Therefore, one cannot satisfy these goals by solely searching in the adjacency matrix space. Indeed, removing a single edge can leave the graph with vastly different connectivity structure and molecular properties. Therefore, we seek an alternative latent space created via neural network transformations that is continuous with respect to graph structure and property.

\begin{figure}[ht]
\begin{center}
\includegraphics[scale=0.35]{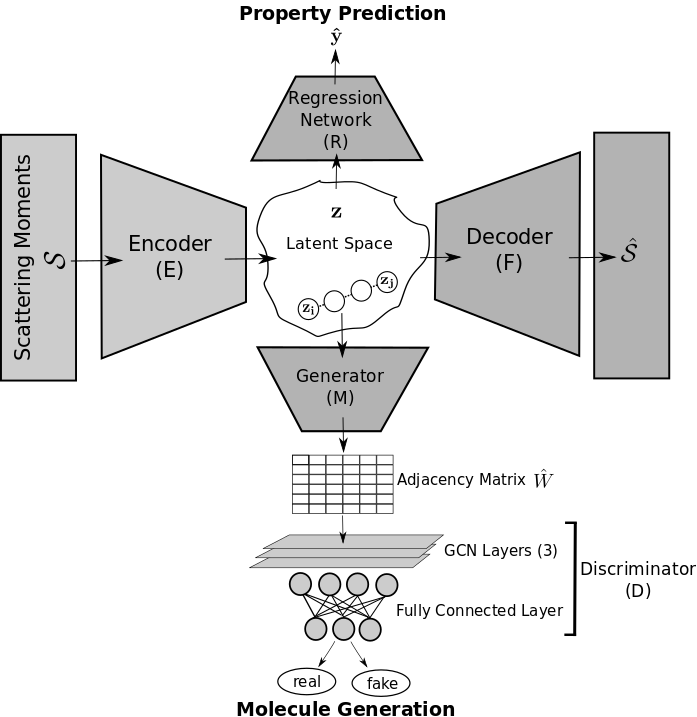}
\end{center}
\caption{Overview of the GRASSY architecture including the 1) geometric scattering moments $\mathbf{S}$ computed on the input graph, passed to an 2) encoder network $E$ whose latent space is regularized for property prediction and for graph generation and trained for reconstruction via the 3) decoder $D$, and the 4) Molecular generation network trained with GAN and interpolation losses.}
\end{figure}

\subsection{Learned scattering moments}
\label{learnedscattering}
Our proposed GRASSY framework uses the graph scattering transform to initially find a high dimensional embedding of the data. For each node-labelled molecular graph, we collect a set of signals associated with each label $\mathbf{x}_{L_1}, \mathbf{x}_{L_2}, \ldots \mathbf{x}_{L_\ell}$. Specifically, we define $\mathbf{x}_{L_i}(v_j)=1$ if the vertex $v_j$ has label $L_i$ and $\mathbf{x}_{L_i}(v_j)=0$ otherwise. We then perform multiscale wavelet transforms on each of these signals using a learnable scattering framework \citep{tong_data-driven_2021}, and collect statistical moments by summing over the vertices (see Eqns. \ref{eqn: scat coeff} and \ref{eqn: scat transform}).  We let $\mathbf{S}=\mathbf{S}(G)$ denote the collection of all scattering moments associated  to a graph $G$ and  use these moments as the input into the regularized autoencoder described below. Importantly, we note that while collecting these statistical moments, we apply a global summation and therefore the dimension of the  the scattering embedding does not depend on the size of the graph.

\subsection{Autoencoder Design}
\label{gsae}

We feed the scattering moments $\mathbf{S}$ into a regularized autoencoder $F(E(X))$, which is penalized by two losses: 

\begin{itemize}
\item The reconstruction loss that penalizes for errors in the reconstruction of scattering moments, i.e., $\mathcal{L}_{r}=\|\mathbf{S}-F(E(\mathbf{S}))\|$, where $E$ is the encoder and $F$ is the decoder.  
\item A regression loss which penalizes the failure of a property prediction network $R$ to  predict a physicochemical properties $p$ of a given molecule from its latent representation, i.e., $\mathcal{L}_{p}=\|p-R(E(\mathbf{S}))\|$.
\end{itemize}

Note that $E(\mathbf{S}(G))$ is our proposed mapping function $f$ that approximates properties described in the problem setup. The molecule generation network, $M$, which we describe in the following subsection, is our proposed pseudo-inversion $f'$ in the problem setup.

\subsection{Molecule Generation}
\label{mol_gen}

If $\mathbf{S}(G_i)$ and $\mathbf{S}(G_j)$ are the scattering moments corresponding to two graphs in  our data set, we let $z_i=E(\mathbf{S}(G_i))$ and $z_j=E(\mathbf{S}(G_i))$ and consider the trajectory  $z_{i \to j}(\alpha) \coloneqq (1-\alpha)z_i + \alpha z_j$. We let $M$ be a multi-layer perceptron which inputs scattering coefficients and outputs an $n\times n$ adjacency matrix and define $\hat{W}_{i \to j}(\alpha)\coloneqq M(z_{i \to j}(\alpha))$. As alluded to earlier, we wish to consider graphs of different sizes, and therefore we will take $n$ to be the size of the largest graph in the data set. Smaller adjacency matrices will be extended to be $n\times n$ via zero padding.

Inspired by the Autoencoder Adversarial Interpolation 
approach applied to images in \citet{oring_autoencoder_2020}, we train $M$ and our discriminator $D$ using three losses adapted to the graph generation setting:

\begin{itemize}
\item The adjacency matrix reconstruction loss, $\mathcal{L}_m = \|W_i^{\text{pad}} - \hat{W}_i \|_F + \| W_j^{\text{pad}} - \hat{W}_j \|_F$, where adjacency matrices $W_i$ and $W_j$ are padded with zero vectors up to size $n \times n$.
\item An adversarial loss, $\mathcal{L}_a = \sum_{k=0}^K -\log(D(\hat{W}_{i \to j}(k/K)))$, where discriminator, $D$, is a graph convolution network (GCN) that outputs a scalar value.
\item A smoothness loss, $\mathcal{L}_s = \sum_{k=0}^K \norm{  \frac{\partial \hat{\mathbf{S}}_{i \to j}(k/K)}{\partial \alpha} }^2$, calculated by taking derivative of scattering moments produced by the decoder, $\hat{\mathbf{S}}_{i \to j}(\alpha) = F(z_{i \to j}(\alpha))$.
\end{itemize}

These losses aid in the production of valid molecular adjacency matrices which have similar structure to nearby points.

\begin{figure}[ht]
\begin{center}
\includegraphics[scale=0.6]{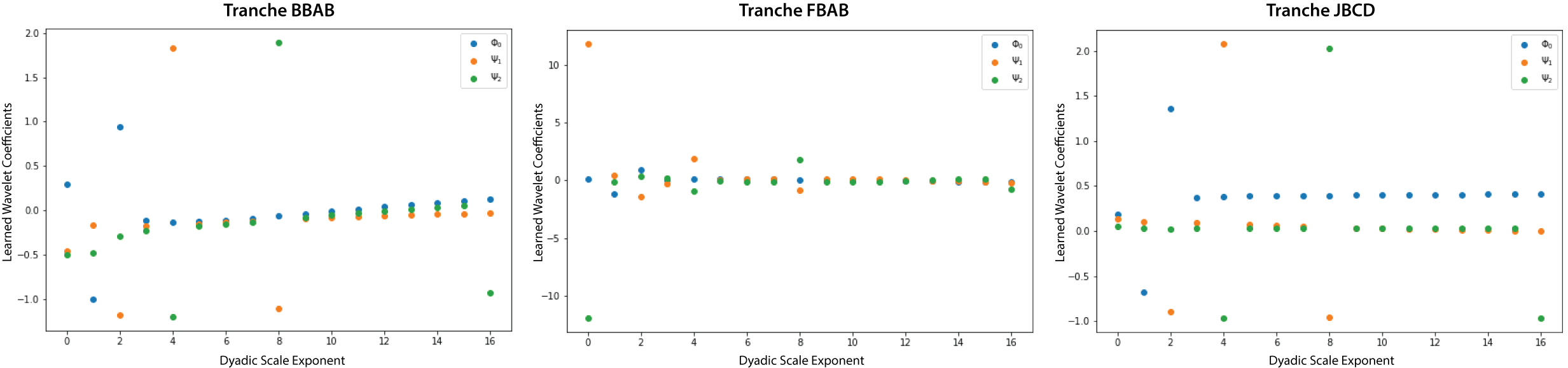}
\end{center}
\caption{Learned geometric scattering wavelet coefficients for 3 ZINC tranches.}\label{fig: LEGS}
\end{figure}

\begin{figure}[ht]
\label{GRASSY_mol_props_ZINC}
\begin{center}
\includegraphics[scale=0.42]{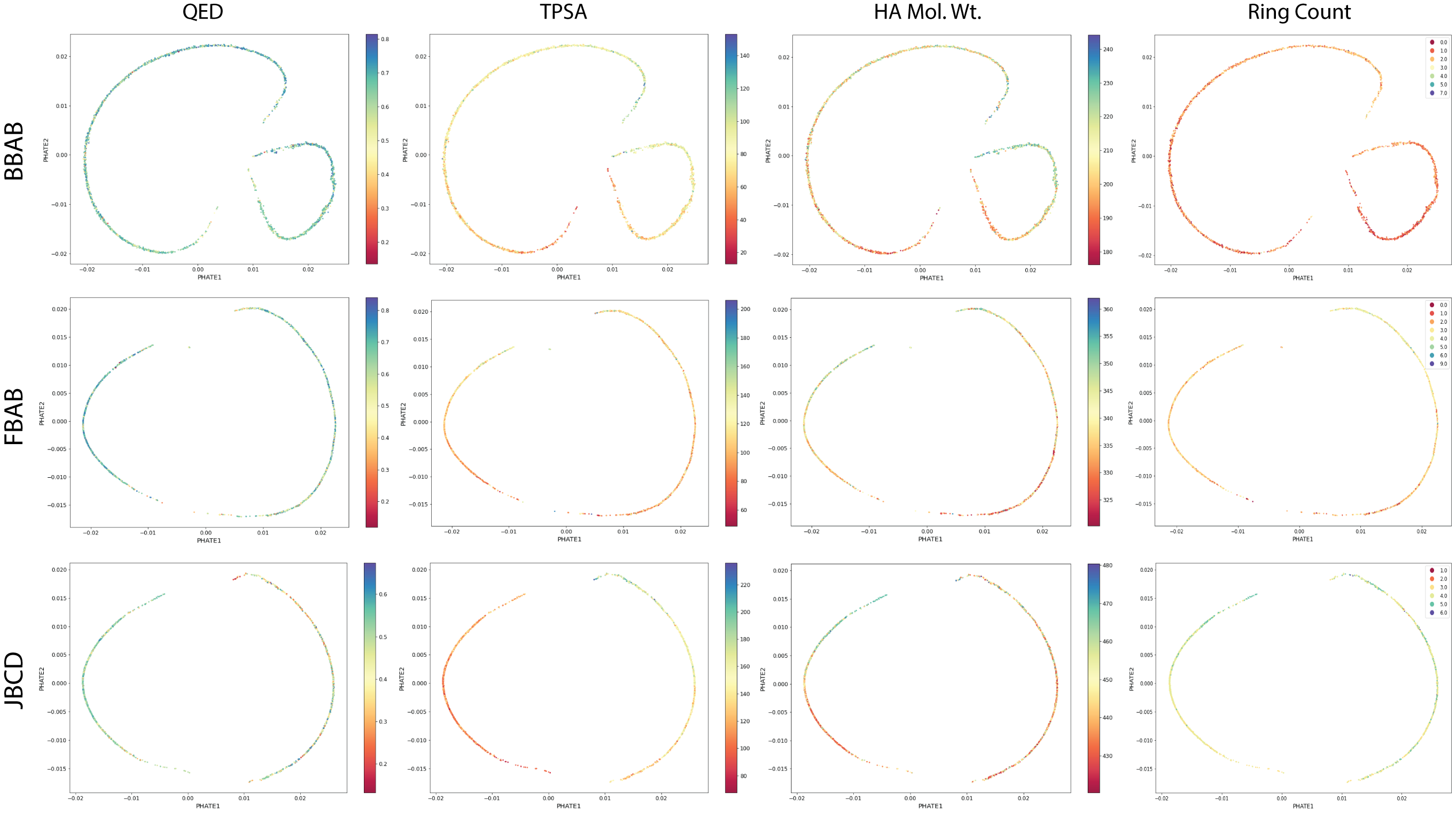}
\end{center}
\caption{Latent representations of molecules in the ZINC dataset, visualized using PHATE, colored by value of physicochemical properties.}\label{fig: Latent rep}
\end{figure}

\begin{figure}[ht]
\label{GRASSY_models_QED}
\begin{center}
\includegraphics[scale=0.52]{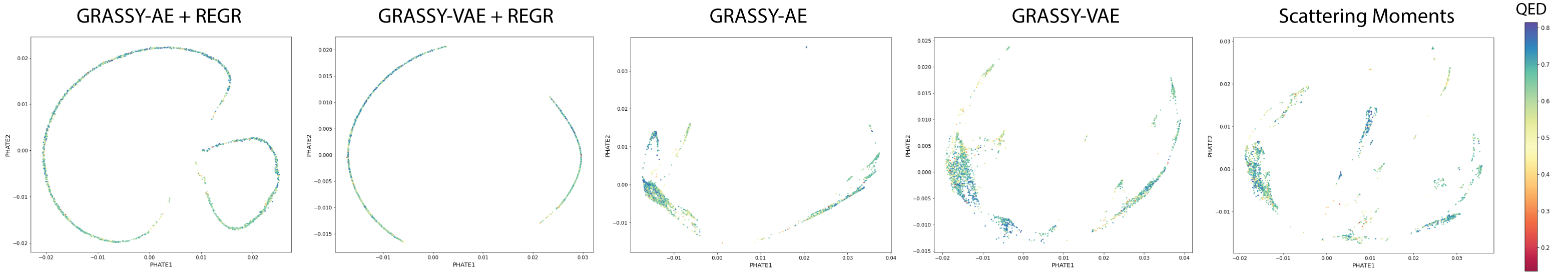}
\end{center}
\caption{Latent representations obtained using ablations of the GRASSY model trained on the BBAB tranche of the ZINC dataset, color-coded by QED for different models.}\label{fig: ablation  trajectories}
\end{figure}

\begin{table}[ht]
{\small
\caption{Molecular property prediction error}
\label{tab: property_prediction_ZINC}
\begin{center}
\begin{tabular}{lllcccc}
\multicolumn{1}{c}{\bf Model} & \multicolumn{1}{c}{\bf Tranche} & \multicolumn{4}{c}{\bf Molecular Properties} \\ 
 &  & QED & TPSA & Mol. Wt. (HA) & \# Rings \\
\hline \\
                & BBAB      & $\mathbf{ 0.1677 \pm 0.13}$ & $\mathbf{ 13.20 \pm 11}$ & $\mathbf{ 22.83 \pm 18}$ & $\mathbf{ 0.6277 \pm 0.52}$ \\
GRASSY-AE + REGR   & FBAB   & 0.2090 $\pm$ 0.16 & 17.15 $\pm$ 15 & $\mathbf{ 32.89 \pm 24}$ & $\mathbf{ 0.7122 \pm 0.57}$ \\
                & JBCD      & $\mathbf{ 0.1941 \pm 0.15}$ & 15.18 $\pm$ 12 & $\mathbf{ 23.30 \pm 22}$ & $\mathbf{ 0.6604 \pm 0.46}$ \\ 
\hline \\
                & BBAB      & 0.2063 $\pm$ 0.16  & 14.91 $\pm$ 13 & 56.67 $\pm$ 21 & 0.6302 $\pm$  0.52\\
GRASSY-VAE + REGR  & FBAB   & $\mathbf{ 0.2041 \pm 0.15}$ & $\mathbf{ 16.94 \pm 14}$ & 36.17 $\pm$ 23 & 0.7130 $\pm$ 0.54 \\
                & JBCD      & 0.2226 $\pm$ 0.17 & $\mathbf{ 14.74 \pm 12}$ & 34.54 $\pm$ 35 & 0.6643 $\pm$ 0.54 \\
\hline \\
                & BBAB      & 0.3947 $\pm$ 0.28 & 78.59 $\pm$ 20 & 210.0 $\pm$ 15 & 1.489 $\pm$ 0.92\\
GRASSY-AE         & FBAB    & 1.401 $\pm$  0.97 & 108.5 $\pm$ 22 & 340.4 $\pm$ 8.6 & 6.528 $\pm$ 1.1 \\
                & JBCD      & 0.4232 $\pm$ 0.26 & 126.9 $\pm$ 26 & 444.5 $\pm$ 14 & 3.152 $\pm$ 0.74 \\
\hline \\
                & BBAB      & 0.7413 $\pm$ 1.3 &  78.53 $\pm$ 20 & 209.4 $\pm$ 15 & 1.364 $\pm$ 0.93 \\
GRASSY-VAE        & FBAB    & 0.9313 $\pm$ 0.33 & 108.5 $\pm$ 22 & 339.5 $\pm$ 8.6 & 3.038 $\pm$ 0.96 \\
                & JBCD      & 2.885 $\pm$ 0.92 & 127.8 $\pm$ 26 &  444.7 $\pm$ 14 & 2.974 $\pm$ 0.86 \\
\hline \\
\end{tabular}
\end{center}
}
\end{table}

\begin{table}[ht]
{\small
\caption{Smoothness of physicochemical properties in latent space}
\label{tab: prop_smoothness_ZINC}
\begin{center}
\begin{tabular}{lllcccc}
\multicolumn{1}{c}{\bf Model} & \multicolumn{1}{c}{\bf Tranche} & \multicolumn{4}{c}{\bf Molecular Properties} \\ 
 &  & QED & TPSA & Mol. Wt. (HA) & \# Rings \\
\hline \\
                & BBAB      & 0.0952 & 0.1716 & 0.0144 & 0.7153 \\
GRASSY-AE + REGR   & FBAB      & 0.1324 & 0.1353 & 0.0021 & 0.3172 \\
                & JBCD      & 0.1516 & 0.0645 & {\bf 0.0028} & 0.0928 \\ 
\hline \\
                & BBAB      &  0.0829 & 0.1785 & {\bf 0.0134}  & 0.6415   \\
GRASSY-VAE + REGR  & FBAB      & 0.1298 & 0.1333 & {\bf 0.0020}  & {\bf 0.3016} \\
                & JBCD      & {\bf 0.1225} & {\bf 0.0541} & 0.0031 & 0.0805 \\
\hline \\
                & BBAB      & {\bf 0.0692}  & 0.1467 & 0.0138 & {\bf 0.6114} \\
GRASSY-AE         & FBAB      & 0.1203  & {\bf 0.1332}  & 0.0023  & 0.3413  \\
                & JBCD      & 0.1383 & 0.0697 & 0.0037 & 0.0837 \\
\hline \\
                & BBAB      & 0.0696 & {\bf 0.1442} & 0.0147 & 0.6266 \\
GRASSY-VAE        & FBAB      & {\bf 0.1201} & 0.1423 & 0.0022 & 0.3593 \\
                & JBCD      & 0.1376 & 0.0713 & 0.0036 & {\bf 0.0800} \\
\hline \\               
\end{tabular}
\end{center}
}
\end{table}

\begin{table}[ht]
{\small
\caption{Fraction of graphs generated with molecule-like structure}
\label{tab: graph_gen_zinc}
\begin{center}
\begin{tabular}{lcccccccc}
\multicolumn{1}{c}{\bf ZINC}  & \multicolumn{2}{c}{\bf \# Atoms} & {\bf Validity} & \multicolumn{4}{c}{\bf Models}\\
{\bf Tranche} & Min. & Max. & {\bf Threshold} & GRASSY & GSAE & GraphAF & MolGAN ($\lambda = 0$) \\ \hline \\
BBAB & 8 & 18 & 5 & {\bf 0.86} & 0.22 & 0.79 & 0.32 \\
FBAB & 16 & 27 & 15 & {\bf 0.94} & 0.17 & 0.76 & 0.46\\
JBCD & 28 & 36 & 25 & {\bf 0.73} & 0.09 & 0.54 & 0.41\\
\hline \\
\end{tabular}
\end{center}
}
\end{table}

\section{Results}
\label{results}

We trained  GRASSY on large datasets of drugs and drug-like molecules from two databases, ZINC \citep{irwin_zinc_2005} and BindingDB \citep{gilson_bindingdb_2016}. ZINC contains drug-like molecules organized into tranches by molecular weight (abbreviated mol. wt.), solubility (logP value), reactivity and commercial availability. BindingDB is a drug-target interaction (DTI) database, containing pairs of molecules and proteins with known binding affinity. We used three tranches from the ZINC database, namely FBAB, BBAB, and JBCD, which are organized by molecular weight. Molecules in BBAB range from 200 to 250 Daltons, molecules in FBAB range from 350 to 375 Daltons, and molecules from JBCD range from 450 to 500 Daltons.  We sampled 2500 molecules from each tranche to use as training data and  used the ChEMBL  database \citep{davies_chembl_2015} to obtain 10 physicochemical for each molecule. These properties include quantitative estimate of drug-likeliness (QED) introduced in \citet{bickerton_quantifying_2012}, total polar surface area (TPSA), general chemical descriptors (mol. wt. and heavy atom mol. wt.), topochemical descriptors (BalabanJ, BertzCT, HallKierAlpha), and Lipinski parameters (number of hydrogen bond donors, number of hydrogen bond acceptors, and ring count). We also trained on known inhibitors of two protein targets on BindingDB, henceforth referred to by their 
UniProtKB identifiers. These targets are P14416, a D(2) dopamine receptor, and P00918, Carbonic anhydrase 2. All BindingDB results are shown in the Appendix.  

To calculate the scattering moments for an individual molecular graph, we labeled each node by atom type and used the signals $\mathbf{x}_{L_1},\ldots,\mathbf{x}_{L_\ell}$ as described in Section \ref{sec: formulation}. We set the number of moments  $Q = 2$ and the number of scales $J = 4$. Since our tranches ranged in molecular weight, we decided to use a learnable version of graph scattering proposed in  \citet{tong_data-driven_2021} to individually learn diffusion scales for each tranche rather than just using dyadic powers $2^j$. The learned wavelet coefficients for each of the three ZINC tranches can be seen in Figure \ref{fig: LEGS}. We then passed the scattering moments into our autoencoder, which applied regression penalties with respect to 10 physicochemical properties in latent space.

We adjusted the learning rates so that our network was properly able to learn the input data on each tranche, and we used an early stopping mechanism that monitored the validation loss to prevent the models from over-fitting our data. Figure \ref{fig: Latent rep} shows the latent space representations of each tranche trained on GRASSY, where the latent space is colored by four different properties, each of which was part of the latent space regression task.

We compare GRASSY to three state-of-the art frameworks for molecular graph generation: GraphAF \citep{shi_graphaf_2020}, MolGAN \citep{de_cao_molgan_2018} and GSAE \citep{castro_uncovering_2020}. For GraphAF, we trained the TorchDrug implementation  using a custom dataloader but without any architectural modifications. For MolGAN, we set the hyperparameter $\lambda$ to $0$ (full RL version). We also compared with experimented with ablations of the GRASSY architecture, turning off and on the regression penalty (with or without REG), as well as toggling variational vs vanilla autoencoder training (GRASSY-VAE vs. GRASSY-AE).

We tested these GRASSY variations according to three criteria:
\begin{itemize}
\item Accuracy of physicochemical prediction in latent space across each of these models, measuring by absolute error. (See Tables \ref{tab: property_prediction_ZINC} and \ref{tab: prop_pred_bindingdb}.)
\item Smoothness of the physicochemical properties in latent space, which we calculated from the graph Laplacian, $L$, of the diffusion potential graph obtained from the PHATE transform of the latent space embeddings. We calculated smoothness by $s \coloneqq \frac{p^TLp}{p^Tp}$, where $p$ is a vector containing properties of the points embedded. (See Tables \ref{tab: prop_smoothness_ZINC} and \ref{tab: prop_smoothness_bindingdb})
\item Fraction of valid molecules generated. (See Table \ref{tab: graph_gen_zinc})
\end{itemize}

Tables \ref{tab: prop_smoothness_ZINC} and \ref{tab: prop_smoothness_bindingdb}
show smoothness, as described above, of physicochemical properties in the latent space of various versions of our architecture. We have made bold values representing the highest smoothness in the latent space for each tranch (or protein target for BindingDB) trained on each property, across the various architectures. We can see that there is no version of our architecture that significantly outperforms the others across the board. Though, in general, the Figure \ref{fig: ablation  trajectories} shows that each version of the GRASSY produces a smooth latent space with respect to the physicochemical properties listed.

Tables \ref{tab: property_prediction_ZINC} and \ref{tab: prop_pred_bindingdb} show the molecular property prediction errors across different versions of our architecture on four different physicochemical properties. On both tables, we made bold the lowest error values for each property, across the various versions of GRASSY. Table \ref{tab: property_prediction_ZINC} shows the prediction error across the three different tranches of ZINC we trained, BBAB, FBAB, and JBCD. We can see that GRASSY-AE + REGR significantly outperforms the other three versions of our architecture with respect to property prediction in the latent space on this dataset. Table \ref{tab: prop_pred_bindingdb} shows that GRASSY-VAE + REGR outperformed our other architectures when trained on the BindingDB data, across protein targets P14416 and P00918.

Molecule-like graphs were created by thresholding the adjacency matrix output of the generator of GRASSY and identifying the largest connected component. The resulting molecule was considered valid if it satisfied the following criteria:
\begin{itemize}
    \item Number of vertices in the graph must exceed a  ``validity threshold", set according to size of molecules in the training tranche. Specifically, single atoms, diatomic and tri-atomic molecules are unlikely to function as drugs or inhibit the activity of any protein target.
    \item Any cycles/rings must not contain more than 10 vertices/atoms. In practice, most drugs contain 5 or fewer rings \citep{aldeghi_two-_2014}.
    \item No vertex should have degree (covalent bonds) exceeding 5. Our data predominantly contains hydrogen, carbon, nitrogen, oxygen, sulphur, chlorine and fluorine atoms, none of which have enough valence electrons to bond with more than 5 atoms simultaneously.  
\end{itemize}

Table \ref{tab: graph_gen_zinc} shows the fraction of valid molecules generated by GRASSY, GSAE \citep{castro_uncovering_2020}, GraphAF \citep{shi_graphaf_2020} and MolGAN\footnote{https://github.com/nicola-decao/MolGAN} (full RL version with hyperparameter $\lambda = 0$ \citep{de_cao_molgan_2018}, according to the criteria above. The minimum and maximum number of atoms in the training molecules are listed to justify the choice of the validity threshold used to mark small generated molecules as invalid.

\section{Conclusions}
\label{conclusions}

We have introduced GRASSY, a novel method for molecule generation. Given a dataset of molecules represented as graphs, we first collect a sequence of scattering moments for each graph. We then train a regularized autoencoder on these scattering moments to produce a latent representation which respects the physicochemical properties of each molecule.   Finally, we produce new molecules by interpolating in latent space an applying a GAN which is trained to produce chemically valid molecules. In our experiments, we see that  our network produces a higher proportion of realistic molecules than other methods. We also see that physicochemical properties of the molecules vary smoothly in our latent space and therefore that our network can be used to predict such properties.


\bibliographystyle{unsrtnat}
\bibliography{references}

\begin{thebibliography}{29}
\providecommand{\natexlab}[1]{#1}
\providecommand{\url}[1]{\texttt{#1}}
\expandafter\ifx\csname urlstyle\endcsname\relax
  \providecommand{\doi}[1]{doi: #1}\else
  \providecommand{\doi}{doi: \begingroup \urlstyle{rm}\Url}\fi

\bibitem[Abu-El-Haija et~al.(2019)Abu-El-Haija, Perozzi, Kapoor, Alipourfard,
  Lerman, Harutyunyan, Ver~Steeg, and Galstyan]{abu2019mixhop}
Sami Abu-El-Haija, Bryan Perozzi, Amol Kapoor, Nazanin Alipourfard, Kristina
  Lerman, Hrayr Harutyunyan, Greg Ver~Steeg, and Aram Galstyan.
\newblock Mixhop: Higher-order graph convolutional architectures via sparsified
  neighborhood mixing.
\newblock In \emph{international conference on machine learning}, pages 21--29.
  PMLR, 2019.

\bibitem[Yun et~al.(2019)Yun, Jeong, Kim, Kang, and Kim]{yun2019graph}
Seongjun Yun, Minbyul Jeong, Raehyun Kim, Jaewoo Kang, and Hyunwoo~J Kim.
\newblock Graph transformer networks.
\newblock \emph{Advances in Neural Information Processing Systems},
  32:\penalty0 11983--11993, 2019.

\bibitem[Mallat(2012)]{mallat2012group}
St{\'e}phane Mallat.
\newblock Group invariant scattering.
\newblock \emph{Communications on Pure and Applied Mathematics}, 65\penalty0
  (10):\penalty0 1331--1398, 2012.

\bibitem[Shi et~al.(2020)Shi, Xu, Zhu, Zhang, Zhang, and
  Tang]{shi_graphaf_2020}
Chence Shi, Minkai Xu, Zhaocheng Zhu, Weinan Zhang, Ming Zhang, and Jian Tang.
\newblock {GraphAF}: a {Flow}-based {Autoregressive} {Model} for {Molecular}
  {Graph} {Generation}.
\newblock \emph{arXiv:2001.09382 [cs, stat]}, February 2020.
\newblock URL \url{http://arxiv.org/abs/2001.09382}.
\newblock arXiv: 2001.09382.

\bibitem[Cao and Kipf(2018)]{de_cao_molgan_2018}
Nicola~De Cao and Thomas Kipf.
\newblock Mol{GAN}: An implicit generative model for small molecular graphs.
\newblock \emph{CoRR}, abs/1805.11973, 2018.
\newblock URL \url{http://arxiv.org/abs/1805.11973}.

\bibitem[Kipf and Welling(2017)]{kipf_semi-supervised_2017}
Thomas~N. Kipf and Max Welling.
\newblock Semi-{Supervised} {Classification} with {Graph} {Convolutional}
  {Networks}.
\newblock \emph{arXiv:1609.02907 [cs, stat]}, February 2017.
\newblock URL \url{http://arxiv.org/abs/1609.02907}.
\newblock arXiv: 1609.02907.

\bibitem[Fan and Huang(2021)]{fan_labeled_2021}
Shuangfei Fan and Bert Huang.
\newblock Labeled {Graph} {Generative} {Adversarial} {Networks}.
\newblock \emph{arXiv:1906.03220 [cs, stat]}, February 2021.
\newblock URL \url{http://arxiv.org/abs/1906.03220}.
\newblock arXiv: 1906.03220.

\bibitem[Gómez-Bombarelli et~al.(2018)Gómez-Bombarelli, Wei, Duvenaud,
  Hernández-Lobato, Sánchez-Lengeling, Sheberla, Aguilera-Iparraguirre,
  Hirzel, Adams, and Aspuru-Guzik]{gomez-bombarelli_automatic_2018}
Rafael Gómez-Bombarelli, Jennifer~N. Wei, David Duvenaud, José~Miguel
  Hernández-Lobato, Benjamín Sánchez-Lengeling, Dennis Sheberla, Jorge
  Aguilera-Iparraguirre, Timothy~D. Hirzel, Ryan~P. Adams, and Alán
  Aspuru-Guzik.
\newblock Automatic {Chemical} {Design} {Using} a {Data}-{Driven} {Continuous}
  {Representation} of {Molecules}.
\newblock \emph{ACS Central Science}, 4\penalty0 (2):\penalty0 268--276,
  February 2018.
\newblock ISSN 2374-7943, 2374-7951.
\newblock \doi{10.1021/acscentsci.7b00572}.
\newblock URL \url{https://pubs.acs.org/doi/10.1021/acscentsci.7b00572}.

\bibitem[Coifman and Maggioni(2006)]{coifman_diffusion_2006}
Ronald~R. Coifman and Mauro Maggioni.
\newblock Diffusion wavelets.
\newblock \emph{Applied and Computational Harmonic Analysis}, 21\penalty0
  (1):\penalty0 53--94, July 2006.
\newblock ISSN 1063-5203.
\newblock \doi{10.1016/j.acha.2006.04.004}.
\newblock URL
  \url{https://www.sciencedirect.com/science/article/pii/S106352030600056X}.

\bibitem[Coifman and Lafon(2006)]{Diffusionmaps}
Ronald~R. Coifman and Stéphane Lafon.
\newblock Diffusion maps.
\newblock \emph{Applied and Computational Harmonic Analysis}, 21\penalty0
  (1):\penalty0 5--30, 2006.
\newblock ISSN 1063-5203.
\newblock \doi{https://doi.org/10.1016/j.acha.2006.04.006}.
\newblock URL
  \url{https://www.sciencedirect.com/science/article/pii/S1063520306000546}.
\newblock Special Issue: Diffusion Maps and Wavelets.

\bibitem[Moon et~al.(2019)Moon, van Dijk, Wang, Gigante, Burkhardt, Chen, Yim,
  van~den Elzen, Hirn, Coifman, Ivanova, Wolf, and Krishnaswamy]{phate}
Kevin~R. Moon, David van Dijk, Zheng Wang, Scott Gigante, Daniel~B. Burkhardt,
  William~S. Chen, Kristina Yim, Antonia van~den Elzen, Matthew~J. Hirn,
  Ronald~R. Coifman, Natalia~B. Ivanova, Guy Wolf, and Smita Krishnaswamy.
\newblock Visualizing structure and transitions in high-dimensional biological
  data.
\newblock \emph{Nature Biotechnology}, 37\penalty0 (12):\penalty0 1482--1492,
  December 2019.

\bibitem[van Dijk et~al.(2017)van Dijk, Nainys, Sharma, Kaithail, Carr, Moon,
  Mazutis, Wolf, Krishnaswamy, and Pe'er]{van2017magic}
David van Dijk, Juozas Nainys, Roshan Sharma, Pooja Kaithail, Ambrose~J Carr,
  Kevin~R Moon, Linas Mazutis, Guy Wolf, Smita Krishnaswamy, and Dana Pe'er.
\newblock Magic: A diffusion-based imputation method reveals gene-gene
  interactions in single-cell rna-sequencing data.
\newblock \emph{BioRxiv}, page 111591, 2017.

\bibitem[Haghverdi et~al.(2016)Haghverdi, B{\"u}ttner, Wolf, Buettner, and
  Theis]{haghverdi2016diffusion}
Laleh Haghverdi, Maren B{\"u}ttner, F~Alexander Wolf, Florian Buettner, and
  Fabian~J Theis.
\newblock Diffusion pseudotime robustly reconstructs lineage branching.
\newblock \emph{Nature methods}, 13\penalty0 (10):\penalty0 845, 2016.

\bibitem[Zou and Lerman(2020)]{zou2020graph}
Dongmian Zou and Gilad Lerman.
\newblock Graph convolutional neural networks via scattering.
\newblock \emph{Applied and Computational Harmonic Analysis}, 49\penalty0
  (3):\penalty0 1046--1074, 2020.

\bibitem[Gama et~al.(2018)Gama, Ribeiro, and Bruna]{gama2018diffusion}
Fernando Gama, Alejandro Ribeiro, and Joan Bruna.
\newblock Diffusion scattering transforms on graphs.
\newblock \emph{arXiv preprint arXiv:1806.08829}, 2018.

\bibitem[Gao et~al.(2019)Gao, Wolf, and Hirn]{pmlr-v97-gao19e}
Feng Gao, Guy Wolf, and Matthew Hirn.
\newblock Geometric scattering for graph data analysis.
\newblock In Kamalika Chaudhuri and Ruslan Salakhutdinov, editors,
  \emph{Proceedings of the 36th International Conference on Machine Learning},
  volume~97 of \emph{Proceedings of Machine Learning Research}, pages
  2122--2131. PMLR, 09--15 Jun 2019.
\newblock URL \url{https://proceedings.mlr.press/v97/gao19e.html}.

\bibitem[Perlmutter et~al.(2019)Perlmutter, Gao, Wolf, and
  Hirn]{perlmutter_understanding_2019}
Michael Perlmutter, Feng Gao, Guy Wolf, and Matthew Hirn.
\newblock Understanding {Graph} {Neural} {Networks} with {Asymmetric}
  {Geometric} {Scattering} {Transforms}.
\newblock \emph{arXiv:1911.06253 [cs, stat]}, November 2019.
\newblock URL \url{http://arxiv.org/abs/1911.06253}.
\newblock arXiv: 1911.06253.

\bibitem[Gama et~al.(2019)Gama, Ribeiro, and Bruna]{gama2019stability}
Fernando Gama, Alejandro Ribeiro, and Joan Bruna.
\newblock Stability of graph scattering transforms.
\newblock \emph{Advances in Neural Information Processing Systems},
  32:\penalty0 8038--8048, 2019.

\bibitem[Min et~al.(2020)Min, Wenkel, and Wolf]{min2020scattering}
Yimeng Min, Frederik Wenkel, and Guy Wolf.
\newblock Scattering {GCN}: Overcoming oversmoothness in graph convolutional
  networks.
\newblock \emph{arXiv preprint arXiv:2003.08414}, 2020.

\bibitem[Min et~al.(2021)Min, Wenkel, and Wolf]{min2021geometric}
Yimeng Min, Frederik Wenkel, and Guy Wolf.
\newblock Geometric scattering attention networks.
\newblock In \emph{ICASSP 2021-2021 IEEE International Conference on Acoustics,
  Speech and Signal Processing (ICASSP)}, pages 8518--8522. IEEE, 2021.

\bibitem[Tong et~al.(2021)Tong, Wenkel, MacDonald, Krishnaswamy, and
  Wolf]{tong_data-driven_2021}
Alexander Tong, Frederik Wenkel, Kincaid MacDonald, Smita Krishnaswamy, and Guy
  Wolf.
\newblock Data-{Driven} {Learning} of {Geometric} {Scattering} {Networks}.
\newblock \emph{arXiv:2010.02415 [cs, stat]}, February 2021.
\newblock URL \url{http://arxiv.org/abs/2010.02415}.
\newblock arXiv: 2010.02415.

\bibitem[Zou and Lerman(2019)]{zou2019encoding}
Dongmian Zou and Gilad Lerman.
\newblock Encoding robust representation for graph generation.
\newblock In \emph{2019 International Joint Conference on Neural Networks
  (IJCNN)}, pages 1--9. IEEE, 2019.

\bibitem[Castro et~al.(2020)Castro, Benz, Tong, Wolf, and
  Krishnaswamy]{castro_uncovering_2020}
Egbert Castro, Andrew Benz, Alexander Tong, Guy Wolf, and Smita Krishnaswamy.
\newblock Uncovering the folding landscape of {RNA} secondary structure using
  deep graph embeddings.
\newblock In \emph{2020 IEEE International Conference on Big Data (Big Data)},
  pages 4519--4528, 2020.
\newblock \doi{10.1109/BigData50022.2020.9378305}.

\bibitem[Oring et~al.(2020)Oring, Yakhini, and Hel-Or]{oring_autoencoder_2020}
Alon Oring, Zohar Yakhini, and Yacov Hel-Or.
\newblock Autoencoder {Image} {Interpolation} by {Shaping} the {Latent}
  {Space}.
\newblock \emph{arXiv:2008.01487 [cs, stat]}, October 2020.
\newblock URL \url{http://arxiv.org/abs/2008.01487}.
\newblock arXiv: 2008.01487.

\bibitem[Irwin and Shoichet(2005)]{irwin_zinc_2005}
John~J. Irwin and Brian~K. Shoichet.
\newblock {ZINC} – {A} {Free} {Database} of {Commercially} {Available}
  {Compounds} for {Virtual} {Screening}.
\newblock \emph{Journal of chemical information and modeling}, 45\penalty0
  (1):\penalty0 177--182, 2005.
\newblock ISSN 1549-9596.
\newblock \doi{10.1021/ci049714}.
\newblock URL \url{https://www.ncbi.nlm.nih.gov/pmc/articles/PMC1360656/}.

\bibitem[Gilson et~al.(2016)Gilson, Liu, Baitaluk, Nicola, Hwang, and
  Chong]{gilson_bindingdb_2016}
Michael~K. Gilson, Tiqing Liu, Michael Baitaluk, George Nicola, Linda Hwang,
  and Jenny Chong.
\newblock {BindingDB} in 2015: {A} public database for medicinal chemistry,
  computational chemistry and systems pharmacology.
\newblock \emph{Nucleic Acids Research}, 44\penalty0 (D1):\penalty0
  D1045--1053, January 2016.
\newblock ISSN 1362-4962.
\newblock \doi{10.1093/nar/gkv1072}.

\bibitem[Davies et~al.(2015)Davies, Nowotka, Papadatos, Dedman, Gaulton,
  Atkinson, Bellis, and Overington]{davies_chembl_2015}
Mark Davies, Michał Nowotka, George Papadatos, Nathan Dedman, Anna Gaulton,
  Francis Atkinson, Louisa Bellis, and John~P. Overington.
\newblock {ChEMBL} web services: streamlining access to drug discovery data and
  utilities.
\newblock \emph{Nucleic Acids Research}, 43\penalty0 (W1):\penalty0 W612--W620,
  July 2015.
\newblock ISSN 0305-1048.
\newblock \doi{10.1093/nar/gkv352}.
\newblock URL \url{https://doi.org/10.1093/nar/gkv352}.

\bibitem[Bickerton et~al.(2012)Bickerton, Paolini, Besnard, Muresan, and
  Hopkins]{bickerton_quantifying_2012}
G.~Richard Bickerton, Gaia~V. Paolini, Jérémy Besnard, Sorel Muresan, and
  Andrew~L. Hopkins.
\newblock Quantifying the chemical beauty of drugs.
\newblock \emph{Nature Chemistry}, 4\penalty0 (2):\penalty0 90--98, February
  2012.
\newblock ISSN 1755-4349.
\newblock \doi{10.1038/nchem.1243}.
\newblock URL \url{https://www.nature.com/articles/nchem.1243}.

\bibitem[Aldeghi et~al.(2014)Aldeghi, Malhotra, Selwood, and
  Chan]{aldeghi_two-_2014}
Matteo Aldeghi, Shipra Malhotra, David~L Selwood, and Ah~Wing~Edith Chan.
\newblock Two- and {Three}-dimensional {Rings} in {Drugs}.
\newblock \emph{Chemical Biology \& Drug Design}, 83\penalty0 (4):\penalty0
  450--461, January 2014.
\newblock ISSN 1747-0277.
\newblock \doi{10.1111/cbdd.12260}.
\newblock URL \url{https://www.ncbi.nlm.nih.gov/pmc/articles/PMC4233953/}.

\end{thebibliography}

\appendix
\section{Appendix}

Figures and tables associated with the BindingDB dataset are included below:

\begin{figure}[h]
\label{GSAE_models_QED_BindingDB}
\begin{center}
\includegraphics[scale=0.5]{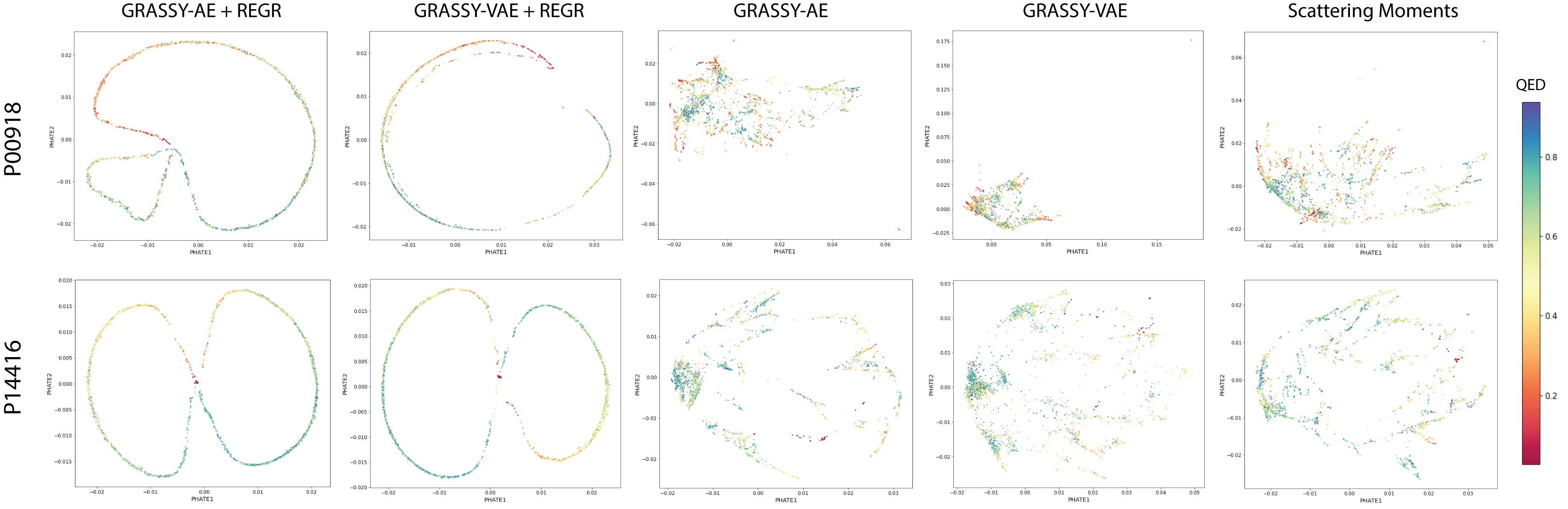}
\end{center}
\caption{Latent representations obtained using ablations of the GRASSY model trained on the BindingDB dataset, color-coded by QED.}
\end{figure}

\begin{table}[h]
{\small
\caption{Accuracy of physicochemical property prediction for BindingDB}
\label{tab: prop_pred_bindingdb}
\begin{center}
\begin{tabular}{llcccc}
\multicolumn{1}{c}{\bf Model} & \multicolumn{1}{c}{\bf Protein} & \multicolumn{4}{c}{\bf Molecular Properties} \\ 
 & {\bf Target}  & QED & TPSA & Mol. Wt. (HA) & \# Rings \\
\hline \\
GRASSY-AE + REGR    & P14416    &  0.3072 $\pm$ 0.43 & 22.52 $\pm$ 30  & $\mathbf{ 49.51 \pm 66}$  & $\mathbf{ 0.8053 \pm 1.6}$  \\
                    & P00918    & $\mathbf{ 0.2911 \pm 0.22}$ & 35.50 $\pm$ 34  & 62.91 $\pm$ 71  & 0.8908 $\pm$ 0.75 \\
\hline \\
GRASSY-VAE + REGR   & P14416  & $\mathbf{ 0.2430 \pm 0.20}$ & $\mathbf{ 18.99 \pm 31}$ & 52.02 $\pm$ 66 &  0.9096 $\pm$ 1.9 \\
                    & P00918  & 0.3395 $\pm$ 0.27  & $\mathbf{ 34.80 \pm 33}$ &  $\mathbf{ 56.72 \pm 64}$  & $\mathbf{ 0.7981 \pm 0.66}$ \\
\hline \\
GRASSY-AE       & P14416    & 0.5672 $\pm$ 0.83 & 56.90 $\pm$ 74 & 399.1 $\pm$ 196  & 3.533 $\pm$ 1.4 \\
                & P00918    & 0.6871 $\pm$ 0.63 & 116.8 $\pm$ 57  & 373.3 $\pm$ 151 & 2.483 $\pm$ 1.5 \\
\hline \\
GRASSY-VAE      & P14416    &  0.7300 $\pm$ 0.35 & 56.86 $\pm$ 74  & 399.7 $\pm$ 197  & 4.154 $\pm$ 1.3  \\
                & P00918    & 0.6001 $\pm$ 0.30 & 117.1 $\pm$ 57 & 372.3 $\pm$ 150  & 2.630 $\pm$ 1.4  \\
\hline \\               
\end{tabular}
\end{center}
}
\end{table}

\begin{table}[h]
{\small
\caption{Smoothness of physicochemical properties in latent space for BindingDB}
\label{tab: prop_smoothness_bindingdb}
\begin{center}
\begin{tabular}{llcccc}
\multicolumn{1}{c}{\bf Model} & \multicolumn{1}{c}{\bf Protein} & \multicolumn{4}{c}{\bf Molecular Properties} \\ 
 & {\bf Target}  & QED & TPSA & Mol. Wt. (HA) & \# Rings \\
\hline \\
GRASSY-AE + REGR   & P14416    & 0.1846 & 0.2813 & 0.0822 & 0.1494 \\
                & P00918    & 0.3786 & 0.4469 & 0.1611 & 0.5016 \\
\hline \\
GRASSY-VAE + REGR  & P14416    & {\bf 0.1746} & 0.3032 & {\bf 0.0716} & {\bf 0.1475} \\
                & P00918    & 0.3345 & 0.3991 & {\bf 0.1294} & 0.3926 \\
\hline \\
GRASSY-AE         & P14416    & 0.1811 & {\bf 0.2385} & 0.0995 & 0.1505 \\
                & P00918    & {\bf 0.2405} & {\bf 0.2436} & 0.1496 & {\bf 0.2954} \\
\hline \\
GRASSY-VAE        & P14416    & 0.2282 & 0.2688 & 0.1148 & 0.1818 \\
                & P00918    & 0.3685 & 0.4035 & 0.2368 & 0.4820 \\
\hline \\               
\end{tabular}
\end{center}
}
\end{table}

\end{document}